\documentclass[journal]{IEEEtran}

\usepackage{cite}
\usepackage{graphicx}
\usepackage{bm}
\usepackage{epsfig}
\usepackage{amsmath}
\usepackage{array}
\usepackage{multirow}
\usepackage{color}
\usepackage{booktabs}
\pdfoutput=1
\usepackage{balance}
\usepackage{amsfonts}
\usepackage{amssymb,bm} 
\usepackage[dvipsnames]{xcolor}
\usepackage{algpseudocode}

\usepackage[ruled,linesnumbered]{algorithm2e}
\usepackage{listings}

\newcolumntype{C}[1]{>{\centering\arraybackslash}m{#1}}
\newcolumntype{R}[1]{>{\raggedleft\arraybackslash}m{#1}}
\newcolumntype{P}[1]{>{\raggedright\arraybackslash}p{#1}}
\newcolumntype{M}[1]{>{\centering\arraybackslash}m{#1}}

\usepackage{url}            

\ifCLASSINFOpdf
\else
\fi

\begin{document}
%
\title{{Decoupled Prototype Learning for Reliable Test-Time Adaptation}}

\author{Guowei Wang, Changxing Ding*, Wentao Tan, Mingkui Tan
\thanks{Guowei Wang is with the School of
Electronic and Information Engineering, South China University of Technology, 381 Wushan Road, Tianhe District, Guangzhou 510000, P.R. China (email: eegw.wang@mail.scut.edu.cn). 

Changxing Ding is with the School of Electronic and Information Engineering, South China University of Technology, 381 Wushan Road, Tianhe District, Guangzhou 510000. P.R. China, and also with the Pazhou Lab, Guangzhou
510330, China (e-mail: chxding@scut.edu.cn). * Corresponding author.

Wentao Tan is with the School of Future Technology, South China University of Technology, 381 Wushan Road, Tianhe District, Guangzhou 510000, P.R. China (e-mail: ftwentaotan@mail.scut.edu.cn).

Mingkui Tan is with the School of Software Engineering and the Key Laboratory of Big Data and Intelligent Robot, Ministry of Education, South China University of Technology, Guangzhou 510006, P.R. China (email: mingkuitan@scut.edu.cn).
}
}

\maketitle

\begin{abstract}

Test-time adaptation (TTA) is a task that continually adapts a pre-trained source model to the target domain during inference. One popular approach involves fine-tuning model with cross-entropy loss according to estimated pseudo-labels.
However, its performance is significantly affected by noisy pseudo-labels. This study reveals that minimizing the classification error of each sample causes the cross-entropy loss's vulnerability to label noise. To address this issue, we propose a novel Decoupled Prototype Learning (DPL) method that features prototype-centric loss computation. First, we decouple the optimization of class prototypes. For each class prototype, we reduce its distance with positive samples and enlarge its distance with negative samples in a contrastive manner. This strategy prevents the model from overfitting to noisy pseudo-labels. Second, we propose a memory-based strategy to enhance DPL's robustness for the small batch sizes often encountered in TTA. 
We update each class's pseudo-feature from a memory in a momentum manner and insert an additional DPL loss. 
Finally, we introduce a consistency regularization-based approach to leverage samples with unconfident pseudo-labels. This approach transfers feature styles of samples with unconfident pseudo-labels to those with confident pseudo-labels. Thus, more reliable samples for TTA are created. The experimental results demonstrate that our methods achieve state-of-the-art performance on domain generalization benchmarks, and reliably improve the performance of self-training-based methods on image corruption benchmarks. The code will be released. 

\end{abstract}


\IEEEpeerreviewmaketitle

\section{Introduction}

\IEEEPARstart {D}{eep} neural networks perform adequately when training and testing data are drawn from the same distribution. However, they often decline significantly in performance when confronted with data from hidden target domains. Test-time adaptation (TTA) \cite{wang2020tent,iwasawa2021test,choi2022improving,chi2021test,chen2022contrastive,wang2022continual,boudiaf2022parameter,niu2022efficient,zhao2022test,TDS} is a recently proposed strategy that adapts a pre-trained source model to the target domain during inference. It concurrently makes predictions for the online data in each batch and utilizes the data to update the model parameters without needing to gather a substantial amount of target data in advance.

\begin{figure}
    \centering
    \includegraphics[width=0.5\textwidth]{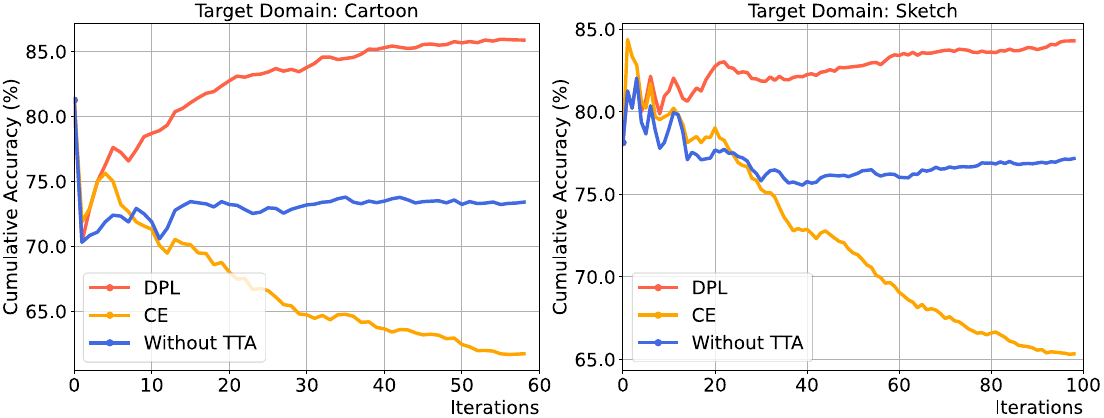}
    \caption{
    Comparisons of test-time cumulative accuracy between the cross-entropy (CE) loss and the decoupled prototype learning (DPL) method during test-time adaptation on the ``cartoon'' domain and the ``sketch'' domain  of the PACS database~\cite{li2017deeper}.
    First, the source model is trained on the remaining source domains using empirical risk minimization. Then, we adapt the source model to the target domain at test time using online data (59 iterations for ``cartoon'' and 99 iterations for ``sketch'').
    }
    \label{fig:fig1}
\end{figure}

    

Despite its convenience, TTA is extremely challenging due to the absence of ground-truth labels. One common approach to addressing this challenge involves generating pseudo-labels and subsequently fine-tuning the model using the cross-entropy (CE) loss and the data with confident pseudo-labels~\cite{lee2013pseudo}.
However, when the domain gap is large, the target domain data's pseudo-label noise increases~\cite{gao2022back, wang2022continual}. As depicted in Fig.~\ref{fig:fig1}, this causes the test-time cumulative accuracy to continually decrease during adaptation. 
Hence, we investigate the underlying causes of CE loss failures and propose a novel loss function to mitigate these issues.

This paper determines that minimizing the classification error of each sample causes the CE loss's vulnerability to label noise.
Specifically, CE loss changes all class prototypes (\textit{i.e.}, classifier weights) simultaneously to fit each sample's pseudo-label. Since the pseudo-labels are noisy, this strategy results in label noise overfitting and backpropagates unreliable gradients to the model's parameters.
Furthermore, without the correction of ground-truth labels, the poorly adapted model would produce more
noisy pseudo-labels, resulting in an unwanted cycle, as illustrated in Fig.~\ref{fig:fig1}.

Therefore,
to overcome this challenge, we propose a novel decoupled prototype learning (DPL) approach for reliable TTA. 
DPL's strength lies in
its prototype-centric loss function, which decouples the optimization of class prototypes.
This strategy emphasizes robust model parameter updates instead of individual pseudo-label fitting. 
First, we make predictions for all the samples in the batch during each iteration and select only those with confident pseudo-labels for loss computation. Then we optimize each prototype independently. For each prototype, our loss function reduces its distance with the positive samples and increases its distance with the negative samples in a contrastive fashion. Finally, we directly use these prototypes as classifier weights to generate pseudo-labels for the incoming data. As a result, the effect of noisy pseudo-labels on model parameter update reduces due to prototype-centric loss computation, enabling the model to continually produce high quality pseudo-lables. This forms a virtuous circle and ultimately enables reliable TTA.

Notably, the above basic DPL form already achieves excellent TTA performance. We further promote its capability from two aspects: first, the batch size during TTA is usually small; thus some classes may lack data based on the pseudo-labels. 
In this case, DPL cannot effectively optimize these classes' prototypes.
To solve this problem, we design a memory that saves pseudo-feature for each class. 
Specifically, the pseudo-feature for one class is updated in a momentum fashion according to online data with pseudo-labels of this category. After that, we construct another DPL loss term to facilitate a more robust TTA using these pseudo-features.


Second, the basic form of DPL utilizes only samples with confident pseudo-labels, disregarding the unconfident ones. We use samples with unconfident pseudo-labels to facilitate DPL's adaptation to large domain shifts. This is because samples with confident pseudo-labels are typically clear in semantics, while the unconfident ones are usually negatively impacted by large domain shifts. As a result, we combine their respective advantages by transferring the styles of the latter to those of the former at the feature-level using adaptive instance normalization (AdaIN)~\cite{huang2017arbitrary}. The new feature labels remain identical to those of the confident samples.
Therefore, our method aligns with the consistency regularization strategy in the semi-supervised learning literature\cite{sohn2020fixmatch}. Finally, this strategy facilities DPL to achieve more reliable and efficient TTA.

To the best of our knowledge, our DPL is the first approach that analyses and manages the negative impact of CE's computation mode in TTA.
We demonstrate that our method achieves state-of-the-art performance through extensive experiments on four domain generalization (DG) benchmarks, (\textit{i.e.}, PACS~\cite{li2017deeper}, VLCS~\cite{fang2013unbiased}, OfficeHome~\cite{venkateswara2017deep}, and TerraIncognita~\cite{beery2018recognition}). This approach also achieves superior performance on single domain generalization (SDG) benchmarks, such as DomainNet-126~\cite{saito2019semi}, and popular image corruption benchmarks, including CIFAR-10-C~\cite{krizhevsky2009learning}, CIFAR-100-C~\cite{krizhevsky2009learning}, and ImageNet-C~\cite{krizhevsky2009learning}.

This paper's remaining sections are structured as follows: Section~\ref{sec:related work} reviews related works on TTA and DG. 
Section~\ref{sec:pre} provides the problem formulation and our motivations.
Section~\ref{sec:methods} explains the details of our DPL approach. Then, we explore the experimental results and extensive analysis in Section~\ref{sec:experiments}. Finally, we conclude in Section~\ref{sec:conclusion}.

\section{Related Works} \label{sec:related work}

This section briefly reviews related works on domain generalization, which directly deploys a generalizable model to the testing data. Then, we elaborate on existing TTA works, which further improves the model's generalization ability during the testing phase.

\subsection{Domain Generalization}
The main goal of domain generalization is to train a model that is capable of generalizing to hidden target domains. Existing DG methods enhance the model's generalization ability from various perspectives, including the learning strategy, data augmentation, and domain-invariant feature extraction.

Popular DG learning strategies include meta-learning~\cite{finn2017model}, self-supervised learning~\cite{jing2020self}, and gradient matching~\cite{yu2020gradient}. Meta-learning approaches~\cite{li2019episodic,zhao2021learning} expose a model to domain shifts during training, enabling it to handle the domain shifts in target domains by separating the training data into meta-train and meta-test sets. Meanwhile, self-supervised learning methods~\cite{carlucci2019domain} allow the model to learn generic features independent of specific downstream tasks, reducing overfitting to domain-specific bias. Other methods~\cite{shi2021gradient,mansilla2021domain,huang2020self} focus on computing robust gradients, assuming the components shared by multiple source domains will be beneficial for training generalizable models.

Furthermore, data augmentation methods~\cite{zhou2020domain,xu2021fourier,li2022uncertainty,zhou2020learning,tan2023style} enhance data diversity to improve model's generalization ability. Depending on the position at which augmentation is applied, these approaches can be divided into feature-~\cite{zhou2020domain,li2022uncertainty,tan2023style}, image-~\cite{zhou2020learning,chen2022mix}, and spectrum-level~\cite{xu2021fourier} methods. For example, MixStyle~\cite{zhou2020domain} augments feature styles by combining the statistics of two image features. 
MiRe~\cite{chen2022mix} merges two images according to their activation maps, which changes their background while remaining their semantic contents. To achieve spectrum-level data augmentation, FACT~\cite{xu2021fourier} combines the low frequencies of two images, enabling the network to concentrate on the images' robust structural information.

The final method category focuses on extracting domain-invariant features, which may be irrelevant to unknown domain shifts during testing. This is often achieved by conducting domain alignment on feature distributions across several source domains, including moment-alignment~\cite{muandet2013domain,erfani2016robust,li2018domain,sun2016deep} and maximum mean discrepancy (MMD)-based alignment~\cite{li2018domaincvpr}. Additionally, certain studies~\cite{zhang2022adaptive, zhang2022towards_disentanglement, jin2021style,niu2023knowledge} assert that domain-specific information may also be helpful. They therefore learn both domain-specific and domain-invariant features from a disentanglement perspective.

Existing DG methods focus on improving performance during training. They often struggle to handle unpredictable domain shifts that occur during testing. This limitation significantly impacts their real-world performance. To address this challenge, TTA was established to study how to enhance the model's generalization ability during the inference stage. 

    

\subsection{Test-Time Adaptation}

TTA methods update a pre-trained source model with each batch of unlabeled data during inference. Unlike  common domain adaptation methods~\cite{yang2021generalized, zhuo2022uncertainty,deng2021informative, wang2022uncertainty}, each online data batch in TTA is only viewed once. Existing TTA methods can be roughly grouped into self-training-,  batch normalization calibration-, 
and prototype-based methods. Moreover, a line of existing research closely related to TTA involves test-time training (TTT)~\cite{sun2020test, liu2021ttt++}. TTT methods usually require elaborately designed auxiliary tasks for the specific downstream tasks~\cite{ni2022meta,liu2022towards,chi2021test} in the model training stage. Thus, this paper focuses on the more flexible and easier-to-use TTA task.

Self-training-based methods optimize model parameters through entropy minimization on the prediction results~\cite{wang2020tent,liang2020we,niu2022efficient,conjugatepl}, or CE loss based on pseudo-labels~\cite{liang2020we, chen2022contrastive, choi2022improving, jang2022test, wang2023feature}. Entropy minimization loss compels the classifier to generate over-confident predictions for the target data, which deteriorates model calibration~\cite{guo2017calibration}. CE loss is vulnerable to noisy labels, which degrades model performance or even results in model collapsing, as illustrated in Fig.~\ref{fig:fig1}.
To relieve this problem, some methods directly reduce the noise in pseudo-labels via various refinement strategies, including augmentation-averaged predictions~\cite{wang2022continual}, shift-agnostic weight regularization~\cite{choi2022improving}, and nearest-neighbor soft voting~\cite{chen2022contrastive}.
Other approaches relieve the impact of noisy pseudo-labels by only optimizing the partial model parameters; however, this strategy affects the model's adaptation capacity~\cite{wang2022continual,chen2022contrastive}. Furthermore, some studies~\cite{niu2022efficient, wang2022continual} recognize that optimization using online data results in catastrophic forgetting, and they alleviate this problem through various techniques, such as Fisher regularization~\cite{niu2022efficient}, stochastic restoration~\cite{wang2022continual}, and weight ensembling~\cite{marsden2023universal}.

Batch normalization (BN) calibration-based methods~\cite{liang2023comprehensive} update the statistics that reflect domain styles in the BN layers. 
For these methods~\cite{singh2019evalnorm,SergeyIoffe_batchrenormalization,burns2021limitations}, accurately estimating BN statistics according to the small amount of online data when facing dramatic domain shifts is challenging. To address this problem, recent studies~\cite{lim2023ttn,you2021test,
hu2021mixnorm,schneider2020improving,hong2023mecta,gong2022note,yuan2023robust, zhao2023delta,zou2022learning, mirza2022norm} calibrate BN statistics by combining those of the source domain with the estimated values from the small online data batch. Very recently, Niu et al.~\cite{niu2023towards} demonstrated that the pre-trained models with group normalization (GN)~\cite{wu2018group} or layer normalization (LN)~\cite{ba2016layer} are more beneficial for the TTA task under small batch sizes.
In this paper, we demonstrate that DPL is insensitive to small batch sizes even when the pre-trained model is equipped with BN.

Prototype-based methods~\cite{iwasawa2021test,jang2022test, wang2023feature,zhang2023adanpc} update a prototype for each class during testing and utilize them to facilitate classification. 
Pioneer methods in this category are typically optimization-free. For example, T3A~\cite{iwasawa2021test} updates prototypes according to a support set for each class. This support set includes each class's original classifier weights and
online data features according to the pseudo-labels. 
Due to pseudo-label noise, the online data features may impair the support set quality. To relieve this problem, Zhang et al.~\cite{zhang2023adanpc} stored the features of numerous source domain data in the support set. Generally, prototype-based methods require representative online data with confident pseudo-labels to update the prototypes. Hence, Wang et al.~\cite{wang2023feature} rectified the target domain's representations according to feature uniformity. Jang et al.~\cite{jang2022test} utilized the
nearest neighbor information to refine the support set.

Our proposed DPL falls under the self-training-based methods. Unlike previous methods that utilize CE loss to optimize model parameters, we discover that CE loss is vulnerable to substantial domain gaps during TTA. In this paper, we explicitly address this issue, enabling reliable TTA with superior performance.

\begin{figure*}[ht]
    \centering
\includegraphics[width=1.0\textwidth]{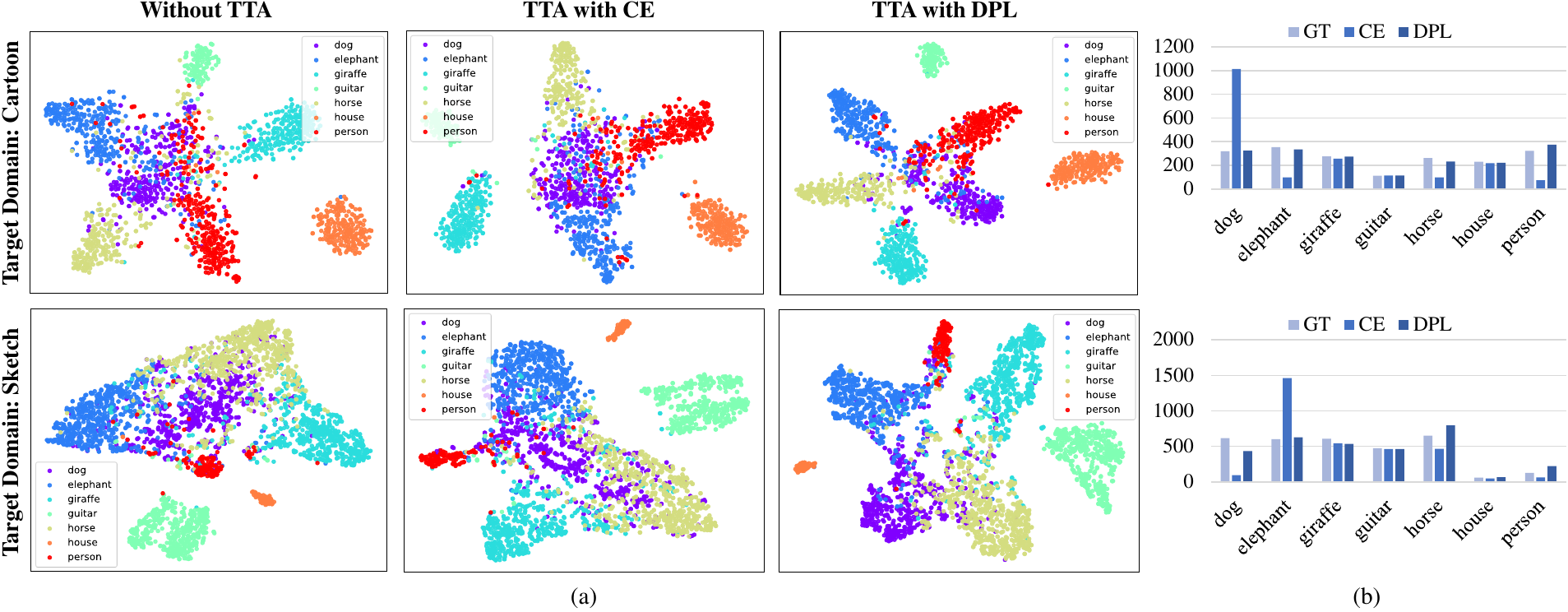}
    \caption{
    (a) T-SNE visualization of feature spaces on the ``cartoon'' (row 1) and ``sketch'' (row 2) domains of the PACS database~\cite{li2017deeper} for the source model, the model optimized by the CE loss, and our DPL, respectively. 
    DPL enables the model to learn disentangled features for the seven categories. (b) The comparison of predictions made by CE and DPL with the ground truth (\textit{i.e.}, ``GT''). 
    The vertical axis presents the number of samples according to the GT or
    pseudo-labels. The horizontal axis represents the categories. The predictions made by DPL are more consistent with GT than those by the CE loss. Best viewed in color.
    }
    \label{fig:fig_methods}
\end{figure*}

\section{Preliminaries}\label{sec:pre}
In this section, we first introduce TTA's problem formulation.
We then present our motivations based on an analysis to standard self-training-based methods.


\textbf{Problem formulation.} Given a model $f_\theta$ pre-trained on the source domain data $\mathcal{D}_\mathcal{S}$, we 
continually feed batches of unlabeled online data stream $\mathcal{X}_n$ from the target domain $\mathcal{D}_\mathcal{T}$ to $f_\theta$. Here $n$ denotes the batch index and
$\mathcal{X}_n=\left\{\textbf{x}_1, \textbf{x}_2, ..., \textbf{x}_i, ...\right\}$.
The purpose of TTA is to utilize these unlabeled online data to adapt parameters of $f_\theta$ to $\mathcal{D}_\mathcal{T}$.

\textbf{Motivations.} The self-training-based methods~\cite{liang2020we,wang2022continual} first generate pseudo-labels for the target domain data as follows:

\begin{equation}\label{eq.1}
\hat{y}_i={\underset{c}{\operatorname{argmax}}(\sigma( f_\theta(\textbf{x}_i)) ^c}),
\end{equation}

\noindent where $f_\theta(\textbf{x}_i)$ includes the predicted probabilities for the target domain data $\textbf{x}_i$, $c$ denotes the class index, $\hat{y}_i$ refers to the assigned class label, $\sigma(\cdot)$ stands for the softmax function, and $\sigma(f_\theta(\textbf{x}_i))^c$ is the \textit{c}-th element of $\sigma(\cdot)$. 

Since the pseudo-labels are noisy, existing approaches usually use a pre-defined threshold $\alpha$ to select data with confident pseudo-labels. Specifically, if the prediction confidence to the pseudo-label $\hat{y}_i$ is above $\alpha$, then this pseudo-label is regarded as confident. We incorporate samples with confident pseudo-labels into the cross-entropy (CE) loss computation, as denoted below:

\begin{equation}\label{eq.2}
    \mathcal{L}_{CE} = -\frac{1}{N} \sum_{i=1}^N \log \frac{\exp \left(\textbf{z}_i \cdot \textbf{w}_{+}\right)}{\sum\nolimits_{\substack{k=1 }}^C \exp\left(\textbf{z}_i \cdot \textbf{w}_{k} \right)},
\end{equation}

\noindent where $N$ denotes the number of samples with confident pseudo-labels, $C$ represents the total number of classes, $\textbf{z}_{i}$ is the feature representation of the \textit{i}-th sample, and ($\cdot$) denotes the dot product. Moreover, $\textbf{w}_{+}$ and $\textbf{w}_{k}$ are the classifier weights of classes $\hat{y}_i$ and class $k$, respectively. For simplicity, we exclude the bias of the classifier weights.

As illustrated in Fig.~\ref{fig:fig_methods}(b), $f_\theta$ tends to make incorrect predictions due to significant domain shifts.
When the CE loss is applied to these noisy pseudo-labels, as outlined in Eq.~\ref{eq.2}, it simultaneously optimizes all $\textbf{w}_k$ $(1\leq k \leq C)$, compelling the model to fit each sample's pseudo-label. 
We refer to this computation manner as the sample-centric mode. 
In this case, this strategy results in unreliable gradients and degrades the model's parameters. Consequently, as shown in Fig.~\ref{fig:fig1}, CE loss results in significant error accumulation and causes the TTA performance collapses. In conclusion, sample-centric computation mode makes CE loss sensitive to label noise in TTA.

\section{Methods on TTA}
\label{sec:methods}

This section presents DPL approach that overcomes noisy pseudo-labels in TTA. Then, two strategies to further improve DPL's performance are proposed.

\subsection{Decoupled Prototype Learning}\label{sub:dpl}
Over-fitting each sample's pseudo-label is inappropriate since pseudo-labels are noisy. Instead, the TTA method should focus on reliably updating the model parameters. Accordingly, we propose a TTA-targeted loss function (\textit{i.e.}, DPL) that computes loss in a prototype-centric manner. Formally,

\begin{small}
\begin{equation}\label{eq.3}
    \mathcal{L}_{DPL^o}\!=\!- \frac{1}{N}\!\sum_{i=1}^N \log \frac{\exp \left( \left \langle \textbf{w}_{+} , \textbf{z}_{i} \right \rangle \!/\tau \right)}{\exp \left( \left \langle \textbf{w}_{+} , \textbf{z}_{i} \right \rangle \!/\tau \right)\!+\!\!\!\!\sum\limits_{\substack{\textbf{z}_{j} \in {\textbf{Z}_{-}}}}  \!\!\!\exp\left(\left \langle \textbf{w}_{+} , \textbf{z}_{j} \right \rangle\!/\tau\right)},
\end{equation}
\end{small}

\noindent where $\textbf{Z}_{-}$ represents a set of feature representations with pseudo-labels that are different from those of $\textbf{z}_{i}$, $\tau$ is the temperature, and $\langle \cdot \rangle$ stands for the cosine similarity. If a single class has more than one sample based on the pseudo-label, a duplicate calculation will occur as these samples share an identical $\textbf{Z}_{-}$. Therefore, we reformulate Eq.~\ref{eq.3} as:

\begin{small}
\begin{equation}\label{eq.4}
    \mathcal{L}_{DPL^{*}}\!=\!- \frac{1}{C'}\!\sum_{k=1}^C \log \frac{\sum\limits_{\textbf{z}_{i} \in {\textbf{Z}^k_{+}}}\!\!\!\exp \left( \left \langle \textbf{w}_{k} , \textbf{z}_{i} \right \rangle \!/\tau \right)}{\sum\limits_{\textbf{z}_{i} \in {\textbf{Z}^{k}_{+}}}\!\!\!\exp \left( \left \langle \textbf{w}_{k} , \textbf{z}_{i} \right \rangle \!/\tau \right)\!+\!\!\!\!\sum\limits_{\substack{\textbf{z}_{j} \notin {\textbf{Z}^{k}_{+}}}}  \!\!\!\exp\left(\left \langle \textbf{w}_{k} , \textbf{z}_{j} \right \rangle\!/\tau\right)},
\end{equation}
\end{small}

\noindent where ${\textbf{Z}^{k}_{+}}$ represents the set of feature representations with the pseudo-label $k$. $C'$ stands for the actual number of categories existing in the current batch. Finally, the prototypes $\textbf{w}_{k}  (1\leq k \leq C)$ represent the source model's classifier weights.

During TTA, the batch size is often small, indicating that some classes may lack data based on the
pseudo-labels. In this case, DPL cannot effectively optimize the prototypes of these classes. To address this issue, we employ a memory bank to generate pseudo-features for these classes.
Specifically, 

\begin{small}
\begin{equation}\label{eq.5}
    \textbf{z}^{*}_{k} \xleftarrow{} \eta \textbf{z}^{*}_{k} + \frac{(1-\eta)}{\left|\textbf{Z}^{k}\right|}  {\sum{\textbf{z}_{i}^{k}}}, 
    {\textbf{z}_{i}^{k}} \in {\textbf{Z}^{k}_{+}},
\end{equation}
\end{small}

\noindent where $\textbf{z}^{*}_{k}$ denotes the pseudo-feature for the \textit{k}-th class, $\eta$ is a momentum coefficient, and $\left| \cdot \right|$ represents cardinality. Moreover, $\textbf{z}^{*}_{k}$ is initialized as the source model's classifier weights. With the obtained pseudo-features, we add a regularization term to DPL$^{*}$:

\begin{small}
    \begin{equation}\label{eq.6}
    \mathcal{L}_{REG}\!=\!- \frac{1}{C}\!\!\sum_{k=1}^{C} \!\log \frac{\exp \left( \left \langle \textbf{w}_{k} ,\! \textbf{z}^{*}_{k} \right \rangle \!/\tau \right)}{\exp \!\left( \left \langle \textbf{w}_{k} ,\!\textbf{z}^{*}_{k} \right \rangle \!/\tau \right)\!+\! \!\sum\limits_{c\neq k} \exp\left(\left \langle \textbf{w}_{k} , \!\textbf{z}^{*}_{c} \right \rangle\!/\tau\right)}.
\end{equation}
\end{small}

Finally, the full version of DPL can be written as follows:

\begin{small}
\begin{equation}\label{eq.7}
     \mathcal{L}_{DPL} =  \mathcal{L}_{DPL^{*}} + \beta * \mathcal{L}_{REG},
\end{equation}
\end{small}

\noindent where $\beta$ is a hyper-parameter that controls the second term's weight.

In DPL, the prototypes are optimized in a decoupled manner. Specifically, we reduce 
the distance of each prototype from its positive samples and increase its distance from negative samples. This differs from the CE loss (Eq.~\ref{eq.2}), which optimizes all class prototypes simultaneously to fit each sample's pseudo-label. As illustrated in Fig.~\ref{fig:fig1} and Fig.~\ref{fig:fig_methods}, our prototype-centric loss function achieves a significantly improved TTA performance.

Specifically, due to significant domain gaps, the source model produces noisy pseudo-labels during the initial TTA stage. Then, the CE loss fits each noisy pseudo-label, magnifying the source model's prediction bias, as revealed in Fig.~\ref{fig:fig_methods}(b). 
Furthermore, it is observed that the features of different classes optimized by the CE loss remain entangled and do not exhibit significant differences compared to the source model, as illustrated in Fig.~\ref{fig:fig_methods}(a). This is because it adopts a sample-centric computation mode, increasing the risk for over-fitting to noisy pseudo-labels in TTA.
In contrast, our proposed DPL loss enables reliable model adaptation to the target domain data. Specifically, DPL succeeds in disentangling different classes' representations. 
These results suggest that DPL facilitates the achievement of reliable TTA.

\subsection{Utilization of Samples with Unconfident Pseudo-Labels}\label{section3.3} 

\begin{figure*}[ht]
    \centering
    \includegraphics[width=0.95\textwidth]{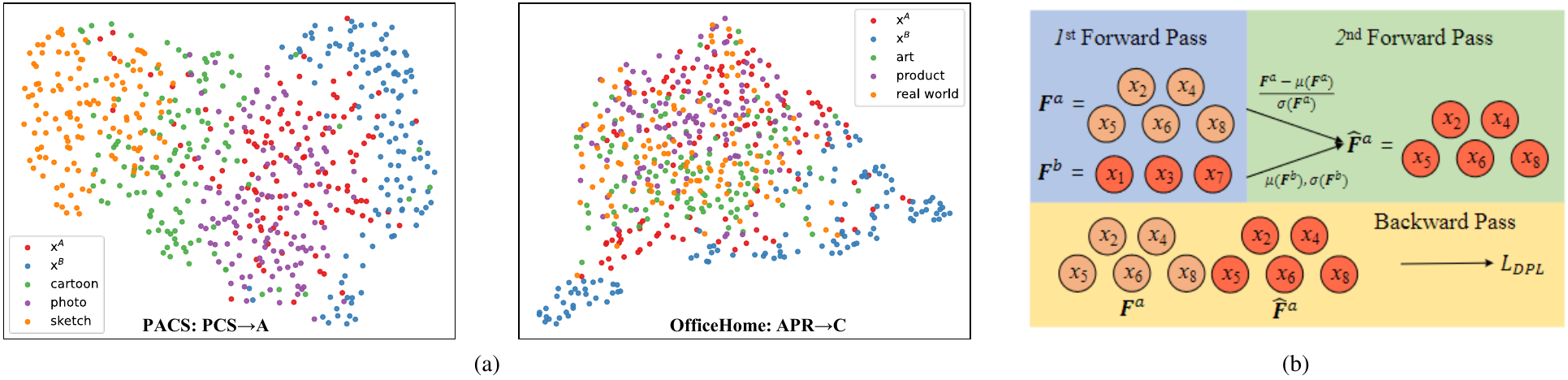}
    \caption{(a) T-SNE visualization of style features (ResNet's stage-1 outputs). Source domain features are colored in \textcolor{orange}{orange}, \textcolor{green}{green}, and \textcolor{violet}{violet}. Target domain features are colored in \textcolor{red}{red} (Confident samples $X^A$) and \textcolor{cyan}{cyan} (Unconfident samples $X^B$). (b) Illustration of the way to utilize samples with unconfident pseudo-labels. We identify samples with confident and unconfident pseudo-labels in the first pass, transfer feature styles from unconfident ones to confident ones in the second pass, and utilize both samples with original and transfer feature styles in our DPL loss.}
    \label{fig_AB}
    \vspace{-0.3cm}
\end{figure*}

In previous sections, we compute loss using only the samples with confident pseudo-labels, while discarding those with unconfident ones. In this subsection, we further enable DPL's adaptability to large domain shifts using samples with unconfident pseudo-labels.

As illustrated in Fig.~\ref{fig_AB} (a), the target domain data's feature styles with confident pseudo-labels are usually similar to those of the source domain data. In contrast, the target domain data with unconfident pseudo-labels exhibit clear feature style differences from those of the source data. 
Therefore, the target domain data with unconfident pseudo-labels may undergo larger domain shifts. Consequently, we utilize the domain information contained in the unconfident samples by transferring their feature styles to the confident ones using AdaIN
~\cite{huang2017arbitrary} at the feature level.

As shown in Fig.~\ref{fig_AB} (b), for each sample with confident pseudo-labels $\textbf{x}^a$, we randomly select a sample with unconfident pseudo-labels $\textbf{x}^b$ from the same batch. The feature maps produced by the selected CNN bottom layer are denoted as $\textbf{F}^a \in \mathbb{R}^{C\times H \times W}$ and $\textbf{F}^b \in \mathbb{R}^{C\times H \times W}$, respectively. Then, we calculate their mean and standard deviation across the spatial dimension within each channel using Eq.~\ref{eq.8} and Eq.~\ref{eq.9}: 

\begin{small}
\begin{equation}\label{eq.8}
    \mu_{c}(\textbf{F})=\frac{1}{H W} \sum_{h=1}^H \sum_{w=1}^W \textbf{F}_{c, h, w},
\end{equation}
\end{small}
\begin{small}
\begin{equation}\label{eq.9}
    \sigma_{c}(\textbf{F})=\sqrt{\frac{1}{H W} \sum_{h=1}^H \sum_{w=1}^W\left(\textbf{F}_{c, h, w}-\mu_{c}(\textbf{F})\right)^2}.
\end{equation}
\end{small}

The obtained vectors $\mu(\textbf{F}^a)$, $\mu(\textbf{F}^b)$, $\sigma(\textbf{F}^a)$, and $\sigma(\textbf{F}^b) \in \mathbb{R}^C$ hold
the style information of $\textbf{x}^a$ and $\textbf{x}^b$. Finally, we transferred the styles of  $\textbf{x}^b$ to $\textbf{x}^a$ using AdaIN~\cite{huang2017arbitrary} and obtained the feature maps for $\textbf{x}^a$ in the new style: 

\begin{small}
\begin{equation}\label{eq.10}
    \hat{\textbf{F}}^{a} = \sigma(\textbf{F}^b) \odot \frac{\textbf{F}^a-\mu(\textbf{F}^a)}{\sigma(\textbf{F}_a)}+\mu(\textbf{F}^b),
\end{equation}
\end{small}

\noindent where $\odot$ denotes element-wise multiplication. The labels of the obtained features remain the same as those of the samples with confident pseudo-labels.

For TTA, we utilize $\textbf{F}^a$ and $\hat{\textbf{F}}^{a}$ for loss computation, with two forward propagations. The first propagation obtains the pseudo-labels and identifies samples with confident and unconfident pseudo-labels, while the second performs style transfer. In this way, we double the size of the confident samples for DPL, enabling the model to efficiently adapt to the new domain.

\textbf{Discussion.} Performing two forward passes in each iteration affects TTA's efficiency. 
An alternative is transferring the feature styles of samples with unconfident pseudo-labels to the next batch of data, which requires these samples to be fed into the model together with the next online data batch.
This strategy achieves feature style transfer and the unconfident sample selection in a single forward pass.
In practice, we observed that this strategy achieves nearly the same performance with noticeably higher efficiency.

\section{The Experiments}
\label{sec:experiments}

In this section, we evaluate our methods' performances with widely-used domain generalization benchmarks, such as PACS~\cite{li2017deeper}, VLCS~\cite{fang2013unbiased}, OfficeHome~\cite{venkateswara2017deep}, TerraIncognita~\cite{beery2018recognition}, and DomainNet-126~\cite{saito2019semi}. We also use image corruption benchmarks, such as CIFAR10-C~\cite{hendrycks2019benchmarking}, CIFAR100-C~\cite{hendrycks2019benchmarking}, and ImageNet-C~\cite{hendrycks2019benchmarking}. Finally, we present comparisons with state-of-the-art methods and detailed ablation studies.

\subsection{Datasets and Evaluation Settings}
\textbf{The datasets.} PACS is composed of 9,991 images in seven categories from four domains, including art, cartoons, photos, and sketches. VLCS consists of 10,729 images in five categories from 4 datasets, including Caltech101~\cite{fei2006one}, LabelMe \cite{russell2008labelme}, SUN09~\cite{choi2010exploiting}, and VOC2007~\cite{everingham2010pascal}. OfficeHome includes 15,588 images in 65 categories from four domains (\textit{i.e.}, art, clipart, product, and real). TerraIncognita~\cite{beery2018recognition} comprises 24,788 images of 10 kinds of wild animals from four domains. DomainNet-126~\cite{saito2019semi} includes 140,006 images with 126 classes from four domains: clipart, painting, real, and sketch.
PACS, TerraIncognita, and DomainNet-126 present significant source-target domain gaps, while VLCS and OfficeHome display relatively smaller ones.

Following the official protocol~\cite{hendrycks2019benchmarking, wang2020tent, wang2022continual}, we apply 15 corruption types (\textit{e.g.}, blur, weather, and digital) with the highest severity level (\textit{i.e.}, level 5) to the clean test images from CIFAR-10/100~\cite{krizhevsky2009learning} and ImageNet~\cite{deng2009imagenet}.
Then, we employ these benchmarks,~\textit{e.g.}, CIFAR-10/100-C~\cite{hendrycks2019benchmarking} and ImageNet-C~\cite{hendrycks2019benchmarking}, to evaluate the robustness of different TTA methods against image corruptions. Moreover, CIFAR-10/100-C and ImageNet-C consist of 10,000 and 50,000 images for evaluation, respectively.

\textbf{The evaluation settings.} 
For the experiments with the DG benchmarks, we adopt the evaluation settings outlined in~\cite{iwasawa2021test}. Specifically, we select a single domain for testing and utilize the remaining domains to pre-train the source model. Then, we evaluate each TTA method's performance on the selected test domain.
For the experiments on single domain generalization (SDG), we select a single domain for training and test the trained model with the remaining three domains. 
Furthermore, for the experiments on corruption benchmarks,
we evaluate a pre-trained model's performance with popular evaluation protocol (\textit{i.e.}, the fully TTA setting~\cite{wang2020tent}). In the fully TTA setting, the model parameters are reset to those of the source model once the domain changes.

\subsection{Implementation Details}
\textbf{The backbone.} 
For the DG benchmark experiments, we mainly adopt
ResNet-50~\cite{he2016deep} as the backbone based on existing studies~\cite{iwasawa2021test,choi2022improving}. We also provide additional experimental results on popular backbones, including ResNet-18~\cite{he2016deep}, Vision Transformer (\textit{e.g.}, ViT-B/16)~\cite{dosovitskiy2020image} and MLP-Mixer (\textit{e.g.}, Mixer-B/16)~\cite{tolstikhin2021mlp}. For DomainNet-126, we directly use the pre-trained ResNet-50~\cite{he2016deep} model provided by~\cite{chen2022contrastive}.

For corruption benchmarks, we adopt the WideResNet-28-10~\cite{zagoruyko2016wide} for CIFAR-10-C~\cite{hendrycks2019benchmarking}, ResNext-29~\cite{xie2017aggregated} for CIFAR-100-C~\cite{hendrycks2019benchmarking}, and ResNet-50~\cite{he2016deep} for ImageNet-C~\cite{hendrycks2019benchmarking} according to existing studies~\cite{wang2022continual,marsden2023universal}. Specifically, the WideResNet-28-10 and ResNext-29 models are provided by the RobustBench library\footnote[1]{https://robustbench.github.io/}, which are pre-trained on CIFAR-10/100 databases~\cite{krizhevsky2009learning}, respectively. Finally, we initialize ResNet-50 using the ImageNet~\cite{deng2009imagenet} pre-trained weights provided by Pytorch\footnote[2]{https://pytorch.org/vision/stable/models.html}.

\textbf{Hyper-parameters and model selection.} 
For experiments on DG benchmarks, we follow the well-established model selection rule in~\cite{gulrajani2020search,iwasawa2021test}.
Specifically, we partition the source domain images into a training and validation sets. These sets consisted of 80\% and 20\% of the source domain images, respectively. Then, we utilize the Adam optimizer~\cite{kingma2014adam} and set the learning rate at 5e-5. 
Finally, the checkpoint that achieves the highest accuracy on the validation set is selected as the pre-trained source model.
During inference, we search the learning rate from \{5e-6, 5e-5, 5e-4\} and iteration steps per adaptation from \{1, 2\}. The batch size is set to 32 and threshold $\alpha$ to 0.9. 
It is important to remember that the hyper-parameters must be selected before accessing the test samples. Furthermore, we determine hyper-parameters on the validation set of the source domains following~\cite{iwasawa2021test}. 
All the experiments are conducted using the T3A library\footnote[3]{https://github.com/matsuolab/T3A} with three seeds. The average results of the three seeds for each experiment are reported, and the entire model's parameters are updated for the experiments on the four DG benchmarks.


Regarding the experiments on the SDG task, we use the Adam optimizer with a learning rate of 4e-5, and the threshold $\alpha$ is fixed to 0.4, which mostly produces acceptable results.
For the corruption benchmarks experiments, we set the hyper-parameters according to those in existing studies~\cite{wang2020tent,wang2022continual}. For example, we use the Adam optimizer with a learning rate of 1e-3 and a batch size of 200 for CIFAR-10/100-C. We also adopt the SGD optimizer with a learning rate of 2.5e-4 and a batch size of 64 for ImageNet-C.
The threshold $\alpha$ is set to 0.9 for CIFAR-10/100-C and 0.4 for ImageNet-C, respectively.
All experiments are conducted on the RMT library\footnote[4]{https://github.com/mariodoebler/test-time-adaptation} with three different seeds. Additionally, we update the affine parameters of the BN layers for experiments on DomainNet-126 and image corruption benchmarks.

\begin{table*}[htp]
\caption{Comparisons of prediction accuracy ($\%$) on domain generalization benchmarks. All experiments were conducted on the T3A library. In this table, * denotes the results reported in the original papers. We implemented the other methods according to their official code. The \textbf{best} and the \underline{second-best} results are highlighted. }

\centering
\setlength\tabcolsep{14pt}
\renewcommand{\arraystretch}{0.8}
\begin{tabular}{l|c|cccc|c}
\toprule
 Method  & Venue & PACS & VLCS & OfficeHome & TerraIncognita  & Avg. \\ \midrule
\midrule
Source &- &82.4 $\pm$ 0.32  &76.0 $\pm$ 0.11  &67.7 $\pm$ 0.22  &47.5 $\pm$ 1.83  &68.4   \\
\midrule
NORM~\cite{li2018adaptive} &- & 83.6 $\pm$ 0.11  &66.4 $\pm$ 0.12    &65.3 $\pm$ 0.32  &38.2 $\pm$ 0.19 &63.4    \\
PL~\cite{lee2013pseudo} &ICML-2013  &82.9 $\pm$ 0.96	&74.5 $\pm$ 0.74	&62.4 $\pm$ 0.68	&42.6 $\pm$ 0.78	&65.6 \\
SHOT~\cite{liang2020we}  &ICML-2020  &84.3 $\pm$ 1.07 &64.7 $\pm$ 2.27   &68.2 $\pm$ 0.15  &33.6 $\pm$ 0.29 &62.7 \\ 
SHOTIM~\cite{liang2020we} &ICML-2020  &84.6 $\pm$ 0.29  &66.9 $\pm$ 0.60    &68.2 $\pm$ 0.05  &33.0 $\pm$ 0.60 &63.2 \\ 
TENT~\cite{wang2020tent} &ICLR-2021  &84.6 $\pm$ 0.31  &72.1 $\pm$ 0.43    &65.7 $\pm$ 0.44  &36.7 $\pm$ 1.55 &64.8 \\ 
\text{T3A}~\cite{iwasawa2021test} &NeurIPS-2021  & 83.4 $\pm$ 0.24 &77.6 $\pm$ 0.80  &69.1 $\pm$ 0.19  & 47.1 $\pm$ 0.48 &69.3 \\
ETA~\cite{niu2022efficient} &ICML-2022  & 83.9 $\pm$ 0.36 &75.2 $\pm$ 0.32  &67.9 $\pm$ 0.03  & \underline{51.6} $\pm$ 0.66 &69.7 \\
SWR\&NSP*~\cite{choi2022improving} &ECCV-2022 &88.9 $\pm$ 0.10  &77.0 $\pm$ 0.50  &\underline{69.2} $\pm$ 0.10 &49.5 $\pm$ 0.80 &\underline{71.3} \\
CPL~\cite{conjugatepl} &NeurIPS-2022  & 82.9 $\pm$ 0.07 &76.0 $\pm$ 0.04  &68.0 $\pm$ 0.01  & 50.3 $\pm$ 0.39 & 69.3  \\
AdaContrast~\cite{chen2022contrastive} &CVPR-2022 &84.7 $\pm$ 0.62  &  74.2 $\pm$ 0.19 & 67.1 $\pm$ 0.21 &48.2 $\pm$ 0.61 &68.6 \\
TSD~\cite{wang2023feature} &CVPR-2023  &\underline{89.4} $\pm$ 0.51	  &74.5 $\pm$ 0.27  & 68.7 $\pm$ 0.14  &37.7 $\pm$ 0.12 &67.5 \\
TAST*~\cite{jang2022test} &ICLR-2023  &81.9 $\pm$ 0.44  &77.3 $\pm$ 0.67  &63.7 $\pm$ 0.52   &42.6 $\pm$ 0.72 &66.4 \\
SAR~\cite{niu2023towards} &ICLR-2023  &84.0 $\pm$ 0.07	&66.8 $\pm$ 0.13	&65.8 $\pm$ 0.17 &39.3 $\pm$ 0.23	&64.0  \\
DomainAdaptor~\cite{zhang2023domainadaptor} &ICCV-2023  & 88.5 $\pm$ 0.16  &\underline{79.6} $\pm$ 0.36   &68.4 $\pm$ 0.03   &39.2 $\pm$ 0.07  &68.9 \\
\midrule
{Ours} &-  &\textbf{89.9} $\pm$ 0.39   &\textbf{81.8} $\pm$ 0.20 &\textbf{69.3} $\pm$ 0.25  &\textbf{52.1} $\pm$ 0.34  &\textbf{73.3}  \\
\bottomrule 
\end{tabular}
\label{table:pacs} 
\vspace{-0.1cm}
\end{table*} 

\begin{table*}[htp]
  \caption{Comparisons in average error rate (\%) on the DomainNet-126 database. The ``clp.,'' ``pnt.,'' ``rel.,'' and ``skt.'' represent the ``clipart,'' ``painting,'' ``real,'' and ``sketch'' domains, respectively. 
  The pretrained ResNet-50 source models were provided by~\cite{chen2022contrastive}. 
  The classification error rate on the source domain is omitted (denoted as ``N/A'').
  The \textbf{best} and the \underline{second-best} methods are highlighted.
  }
  \label{table:domainnet}
  \setlength\tabcolsep{4pt}
 \resizebox{\linewidth}{!}{
 \renewcommand{\arraystretch}{0.5}
 {
 \centering
 
 \begin{tabular}{c|c c c c c  || c | c c c c c  || c |c c c c c }
  \toprule[0.6pt]
   Source 
   & \it{clp.}  & \it{pnt.}  & \it{rel.} & \it{skt.} & Avg.err 
   & Norm~\cite{li2018adaptive} & \it{clp.}  & \it{pnt.}  & \it{rel.} & \it{skt.} & Avg.err 
   & TENT~\cite{wang2020tent} & \it{clp.}  & \it{pnt.}  & \it{rel.} & \it{skt.} & Avg.err \\
  \midrule
  
  \it{clp.} &N/A  &55.3  &40.1  &52.5 &49.3
  &\it{clp.} &N/A  &51.2  &36.9  &53.4  &47.2   
  &\it{clp.} &N/A  &49.2  &38.0  &49.7 &46.6     
  \\
  
  \it{pnt.} &46.7  &N/A  &24.7  &53.8  &41.7   
  & \it{pnt.} &47.2 &N/A  &26.4  &49.1  &40.9   
  & \it{pnt.} &44.2  &N/A  &26.0  &45.0  &38.4   \\

  \it{rel.} &44.9  &37.4  &N/A  &53.4  &45.2   
  & \it{rel.} &47.3  &38.2  &N/A  &53.6  &46.4   
  & \it{rel.} &44.1  &35.3  &N/A  &48.2 &42.5   \\
  
  \it{skt.} &44.8  &49.2  &40.4  &N/A  &44.8   
  & \it{skt.} &42.2  &42.3  &33.7  &N/A &39.4   
  & \it{skt.} &39.9  &42.3  &1.8  &N/A  &39.2   \\

  &- &-&-&-&45.3(\textcolor{blue}{-})
  &&-&-&-&-&43.5(\textcolor{blue}{$\downarrow$1.8})
  &&-&-&-&-&41.7(\textcolor{blue}{$\downarrow$3.6}) 
  \\
  \midrule \midrule
  
  SHOT~\cite{liang2020we} & \it{clp.}  & \it{pnt.}  & \it{rel.} & \it{skt.} & Avg.err 
  
   & SHOTIM~\cite{liang2020we} & \it{clp.}  & \it{pnt.}  & \it{rel.} & \it{skt.} & Avg.err 
   
   & PL~\cite{lee2013pseudo} & \it{clp.}  & \it{pnt.}  & \it{rel.} & \it{skt.} & Avg.err \\
  \midrule
  
  \it{clp.} &N/A  &48.7  &36.0  &49.3  & 44.7  
  & \it{clp.} &N/A  &48.8  &36.0  &49.4  &44.7   
  & \it{clp.} &N/A  &49.0  &35.7  &49.7  &44.8  \\

  \it{pnt.} &44.6  &N/A  &25.4  &45.0  &38.3   
  & \it{pnt.} &44.7 &N/A  &25.4  &45.2  &38.4   
  & \it{pnt.} &44.7  &N/A  &25.3  &45.7  &38.7   \\

  \it{rel.} &44.4  &35.5  &N/A  &48.6  &42.8   
  & \it{rel.} &44.5  &35.6  &N/A  &48.8  &43.0   
  & \it{rel.} &44.7  &35.9  &N/A &48.9 &43.2   \\
  
  \it{skt.} &40.0  &40.7  &33.1  &N/A  & 37.9  
  & \it{skt.} &40.0  &40.7  &33.1  &N/A  & 37.9  
  & \it{skt.} &40.4  &41.0  &32.9  &N/A &38.1   \\

  &- &-&-&-&40.9(\textcolor{blue}{$\downarrow$4.4}) 
  &&-&-&-&-&41.0(\textcolor{blue}{$\downarrow$4.3}) 
  &&-&-&-&-&41.2 (\textcolor{blue}{$\downarrow$4.1}) 
  \\
   \midrule \midrule
    
    SAR~\cite{niu2023towards} 
   & \it{clp.}  & \it{pnt.}  & \it{rel.} & \it{skt.} & Avg.err 
   & CPL~\cite{conjugatepl} & \it{clp.}  & \it{pnt.}  & \it{rel.} & \it{skt.} & Avg.err 
   & AdaContrast~\cite{chen2022contrastive} & \it{clp.}  & \it{pnt.}  & \it{rel.} & \it{skt.} & Avg.err \\
  \midrule
  
  \it{clp.} &N/A  &49.6  &36.2  &49.4  &45.1   
  & \it{clp.} &N/A  &50.1  &41.5  &50.6  &47.4   
  & \it{clp.} &N/A  &45.6  &29.1  &47.6  &\underline{40.8}   \\

  \it{pnt.} &44.7  &N/A  &25.9  &45.1  &38.6  
  & \it{pnt.} &45.3 &N/A  &29.8  &46.3  &40.5  
  & \it{pnt.} &41.3  &N/A  &22.7  &43.4  &35.8   \\

  \it{rel.} &45.5  &36.9  &N/A  &50.0  &44.1  
  & \it{rel.} &46.2  &36.9  &N/A  &51.2  &44.8   
  & \it{rel.} &41.3  &34.7  &N/A  &47.2 &41.1   \\
  
  \it{skt.} &40.2  &41.0  &33.4  &N/A  &38.2  
  & \it{skt.} &41.1  &42.2  &39.7  &N/A  & 41.0  
  & \it{skt.} &37.2  &38.2  &27.1  &N/A  &\underline{34.2}   \\

   &- &-&-&-&41.5(\textcolor{blue}{$\downarrow$3.8}) 
  &&-&-&-&-&43.4(\textcolor{blue}{$\downarrow$1.9}) 
  &&-&-&-&-&38.0(\textcolor{blue}{$\downarrow$7.3}) 
  \\
  \midrule \midrule
   Cotta~\cite{wang2022continual} 
   & \it{clp.}  & \it{pnt.}  & \it{rel.} & \it{skt.} & Avg.err 
   & Rotta~\cite{yuan2023robust} & \it{clp.}  & \it{pnt.}  & \it{rel.} & \it{skt.} & Avg.err 
   & RMT~\cite{dobler2023robust}  & \it{clp.}  & \it{pnt.}  & \it{rel.} & \it{skt.} & Avg.err \\
  \midrule
  
  \it{clp.} &N/A  &44.7  &30.9  &47.2  &40.9   
  & \it{clp.} &N/A  &50.2  &35.4  &51.9  &45.8   
  & \it{clp.} &N/A  &47.9  &38.0  &45.7  &43.9   \\

  \it{pnt.} &40.6  &N/A  &24.0  &42.8  &35.8  
  & \it{pnt.} &46.3  &N/A  &25.1  &48.2  & 39.9  
  & \it{pnt.} &47.9 &N/A  &35.0  &49.3  &44.1 
  \\

  \it{rel.} &42.3  &33.3  &N/A  &47.4  &41.0  
  & \it{rel.} &45.8  &37.1  &N/A  &53.1  & 45.3  
  & \it{rel.}&44.6  &38.2  &N/A  &49.1  &44.0   \\
  
  \it{skt.} &36.3  &37.6  &29.9  &N/A  &34.6  
  & \it{skt.} &41.3  &41.6  &32.6  &N/A  &38.5  
  & \it{skt.} &39.4  &42.4  &37.7  &N/A  &39.8   \\
  &- &-&-&-&38.1(\textcolor{blue}{$\downarrow$7.2}) 
  &&-&-&-&-&42.4(\textcolor{blue}{$\downarrow$2.9}) 
  &&-&-&-&-&43.0(\textcolor{blue}{$\downarrow$2.3})
  \\
  \midrule \midrule
    ROID~\cite{marsden2023universal}
   & \it{clp.}  & \it{pnt.}  & \it{rel.} & \it{skt.} & Avg.err 
   & Ours & \it{clp.}  & \it{pnt.}  & \it{rel.} & \it{skt.} & Avg.err 
   & AdaContrast~\cite{chen2022contrastive}+Ours  & \it{clp.}  & \it{pnt.}  & \it{rel.} & \it{skt.} & Avg.err \\
  \midrule
  
  \it{clp.} &N/A  &47.5  &35.7  &48.0  &43.7   
  & \it{clp.} &N/A  &47.4  &35.7  &46.4  &43.2   
  & \it{clp.} &N/A  &43.9  &28.4  &47.2  &\textbf{39.9}   \\

  \it{pnt.} &40.9  &N/A  &25.2  &42.9  &36.3  
  & \it{pnt.} &40.0  &N/A  &23.7  &40.2  &\textbf{34.6}  
  & \it{pnt.} &41.3 &N/A  &21.4  &42.8  &\underline{35.2}   \\

  \it{rel.} &41.5  &34.8  &N/A  &45.8 &40.7 
  & \it{rel.} &39.7  &33.2  &N/A  &44.6  &\textbf{39.2}  
  & \it{rel.} &41.3  &33.8  &N/A  &47.5  &\underline{40.8}   \\
  
  \it{skt.} &39.5  &40.0  &32.9  &N/A  &37.5 
  & \it{skt.} &37.3  &39.5  &32.5  &N/A  &36.4   
  & \it{skt.} &36.0  &36.7  &26.5  &N/A  &\textbf{33.1}   \\
   &- &-&-&-& 40.0(\textcolor{blue}{$\downarrow$5.3})
  &&-&-&-&-&38.4(\textcolor{blue}{$\downarrow$6.9}) 
  &&-&-&-&-&\textbf{37.3}(\textcolor{blue}{$\downarrow$8.0}) 
  \\
 
 \bottomrule[0.6pt]
 \end{tabular}\label{domainnet-126}
 }
 }
\end{table*}

\subsection{Comparisons with State-of-the-Art Methods}

\textbf{The baselines.} We compare our approach's performance with three existing categories of TTA methods: (1) the self-training-based methods, including \textbf{\textit{PL}}~\cite{lee2013pseudo}, \textbf{\textit{SHOT}}~\cite{liang2020we}, 
\textbf{\textit{SHOT-IM}}~\cite{liang2020we},
\textbf{\textit{TENT}}~\cite{wang2020tent},
\textbf{\textit{EATA}}~\cite{niu2022efficient},
\textbf{\textit{SWR\&NSP}}~\cite{choi2022improving},
\textbf{\textit{CPL}}~\cite{conjugatepl},
\textbf{\textit{AdaContrast}}~\cite{chen2022contrastive},
\textbf{\textit{Cotta}}~\cite{wang2022continual},
\textbf{\textit{Rotta}}~\cite{yuan2023robust}, 
and \textbf{\textit{SAR}}~\cite{niu2023towards}; (2) the prototype-based methods, including \textbf{\textit{T3A}}~\cite{iwasawa2021test}, \textbf{\textit{TAST}}~\cite{jang2022test}, and  \textbf{\textit{TSD}}~\cite{wang2023feature}; and (3) the BN calibration-based methods, including \textbf{\textit{NORM}}~\cite{li2018adaptive}. 
To facilitate a fair comparison, we consistently use ResNet-50 as the backbone. Note that SWR\&NSP~\cite{choi2022improving} and TSD~\cite{wang2023feature}
employ a batch size of 128, while all the other methods adopt a smaller batch size of 32. These settings are consistent with those in the original papers~\cite{iwasawa2021test,jang2022test}. Furthermore, we compare ETA~\cite{niu2022efficient} (\textit{i.e.}, EATA without Fisher regularization) instead of EATA~\cite{niu2022efficient} in Table~\ref{table:pacs}, as EATA assumes the availability of source domain data.

\begin{table*}[htp]

\caption{Comparisons in average error rate (\%) on ImageNet-C. The \textbf{best} and the \underline{second-best} results are highlighted. 
} 
    \centering
    
        \setlength\tabcolsep{4pt}
        \renewcommand{\arraystretch}{0.5}
        
      \scriptsize

        \begin{tabular}{ll|c|ccccccccccccccc}
        \toprule
        \rule{0pt}{5pt} & Methods & Avg. err & Gaus. & Shot. & Impu. & Defo. & Glas. & Moti. & Zoom. & Snow & Fros. & \, Fog \, & Brig. & Cont. & Elas. & Pixe. & Jpeg. \\ \midrule \midrule
       
         & SOURCE~  
         &82.0
         &97.82 &97.10 &98.10 &81.70 &89.76 &85.18 &77.94 &83.46 
         &77.06 &75.90 &41.34 &94.54 
         &82.54 &79.32 &68.56
         \\
        & NORM~\cite{li2018adaptive} 
         &68.6
         &85.04 &84.20 &85.04 &85.14 &84.38
         &73.18 &61.22 &65.86 &68.12 &51.98
         &34.82 &83.08 &55.94 &51.34 &59.86
        \\
         & PL~\cite{lee2013pseudo} 
         &67.0
         &82.50	&81.78	&83.20	&83.24	&82.42	&71.58	&59.98	&64.26	&66.88	&50.36	&34.62	&82.50	&54.90	&50.20	&57.84
         \\
          & SHOT~\cite{liang2020we} 
         &66.1
         &82.06	&80.78	&82.58	&82.36	&81.88	&70.72	&59.08	&63.16	&66.18	&49.56	&34.50	&79.36	&53.94	&49.12	&56.30
         \\

        & TENT~\cite{wang2020tent}
        &65.3
         &81.40 &79.98 &81.98 &81.70 &81.50
         &69.72 &58.04 &62.04 &65.58 &48.72
         &34.16 &78.78 &52.88 &48.00 &55.26
        \\
          & SAR~\cite{niu2023towards}
       &65.2
         &81.98 &80.94 &81.06 &81.24 &80.98
         &69.38 &57.78 &61.88 &65.54 &49.06
         &34.40 &77.02 &53.48 &48.08 &55.74
          \\
         
           & AdaContrast~\cite{chen2022contrastive}
        &69.2
         &82.86 &83.16 &82.46 &86.18 &83.92
         &76.32 &63.06 &64.74 &68.42 &52.02
         &35.96 &88.12 &56.74 &53.68 &60.62
       
        \\
         & CPL~\cite{conjugatepl} 
        &65.9
         &83.46 &82.62 &83.28 &83.52 &82.58
         &70.56 &58.10 &62.46 &65.80 &47.92
         &33.88 &79.60 &52.34 &47.20 &55.66
        \\
          & CoTTA~\cite{wang2022continual} 
          &68.2
         &84.68 &83.96 &84.72 &86.34 &83.74
         &72.18 &60.42 &65.12 &67.64 &51.40
         &34.58 &82.76 &55.62 &50.36 &59.14
            \\
            & RoTTA~\cite{yuan2023robust}
            &69.6
         &88.26 &85.56 &88.76 &84.92 &83.98
         &74.52 &61.78 &67.44 &70.12 &53.28
         &34.62 &83.86 &55.68 &51.52 &59.62
         \\
          & TSD~\cite{wang2023feature}
         &68.3
        &84.52 &84.08 &84.58 &84.80 &83.60
         &72.56 &60.76 &64.98 &67.70 &51.28
         &34.90 &85.86 &55.48 &50.24 &58.78
         \\
          & RMT~\cite{dobler2023robust}
         &68.2
         &77.12 &76.22 &76.42 &79.14 &79.82
         &70.24 &64.72 &65.84 &70.12 &56.90
         &49.40 &78.10 &59.06 &57.04 &62.08
         \\
          & Ours
         &64.7
         &80.94 &82.06 &79.68 &76.16 &77.64
         &70.44 &60.44 &64.16 &64.08 &49.54
         &34.22 &72.70 &54.10 &48.20 &55.84
         \\ \cmidrule{2-18}
        
       & SHOTIM~\cite{liang2020we}
       &66.2
        &82.36 &80.94 &82.68 &82.62 &82.00
         &70.90 &59.12 &63.42 &66.28 &49.54
         &34.56 &79.42 &53.96 &49.20 &56.46
         \\
          & + Ours
          &59.3(\textcolor{blue}{$\downarrow$6.9})
         &74.58 &72.80 &74.02 &70.12 &70.70
         &64.12 &56.26 &57.86 &59.78 &46.52
         &34.22 &65.04 &49.94 &44.76 &48.88
        \\ \cmidrule{2-18}
        & EATA~\cite{niu2022efficient} 
        &60.5
         &75.50 &74.26 &76.50 &76.72 &76.14
         &63.20 &53.84 &56.60 &61.54 &44.86
         &33.32 &69.92 &48.80 &44.56 &51.22
        
         \\
          & + Ours
         &58.0(\textcolor{blue}{$\downarrow$2.5})
         &73.18	&71.02	&73.02	&73.28	&73.24	&59.08	&52.00	&53.56	&59.88	&43.04	&33.40	&67.18	&46.08	&42.50	&48.82
         \\ \cmidrule{2-18}
           & ROID~\cite{marsden2023universal}
           &\underline{57.3}
        &71.84 &70.28 &72.00 &73.58 
        &73.02 &59.38 &51.64 &52.96 
        &59.48 &42.86 &33.04 &63.22 &45.54 &42.50 &48.52 
        \\
         & + Ours
         &\textbf{56.2}(\textcolor{blue}{$\downarrow$1.1})
         &70.46 &67.88 &69.70 &71.66 &72.00
         &57.66 &50.94 &51.34 &58.50 &42.64
         &33.78 &63.00 &44.74 &41.72 &47.30
         \\
           \bottomrule
    \end{tabular}\label{tta.reset.imagenetc}

\end{table*}

\begin{table*}[]
\caption{Comparisons in average error rate (\%) for all five corruption levels on CIFAR-10-C, CIFAR-100-C, and ImageNet-C datasets. The \textbf{best} and the \underline{second-best} results are highlighted. 
} 
 \centering
    \setlength\tabcolsep{4pt}
\renewcommand{\arraystretch}{0.5}
    
    \scriptsize
    \begin{tabular}{l|cccccc|cccccc|cccccc}
    \toprule
        \rule{0pt}{5pt}
     Dataset& \multicolumn{6}{c|}{CIFAR-10-C} & \multicolumn{6}{c|}{CIFAR-100-C}  & \multicolumn{6}{c}{ImageNet-C}                                                                                         \\ \midrule
\begin{tabular}[c]{@{}l@{}}Corruption\\ Level\end{tabular} 

& \multicolumn{1}{c}{@1} & \multicolumn{1}{c}{@2} & \multicolumn{1}{c}{@3} & \multicolumn{1}{c}{@4} & \multicolumn{1}{c}{@5} & \multicolumn{1}{c|}{Avg.err} & \multicolumn{1}{c}{@1} & \multicolumn{1}{c}{@2} & \multicolumn{1}{c}{@3} & \multicolumn{1}{c}{@4} & \multicolumn{1}{c}{@5} & \multicolumn{1}{c|}{Avg.err} & \multicolumn{1}{c}{@1} & \multicolumn{1}{c}{@2} & \multicolumn{1}{c}{@3} & \multicolumn{1}{c}{@4} & \multicolumn{1}{c}{@5} & \multicolumn{1}{c}{Avg.err} \\ \midrule \midrule

Source 
&13.1 &18.7   &25.1  &32.4  &43.5  &26.6
&26.2 &29.6  &33.2  &38.9  &46.5   &34.9 
&39.3   &50.1 &60.2 &72.2   &82.0  &60.8 \\ 
Norm~\cite{li2018adaptive}                                  &9.5 &11.6   &13.5  &16.6  &20.4  &14.3
&26.6 &28.1  &29.9  &32.2 &35.5   &30.5 
&34.1   &41.9 &48.5 &58.4   &68.6  &50.3                    \\ 
PL~\cite{lee2013pseudo} 
&9.2 &10.9   &12.7  &15.2  &19.5  &13.5
&24.5 &25.8  &27.2  &29.2 &31.8   &27.7
&33.8   &41.2 &47.5 &56.9   &67.0  &49.3
\\ 
SHOT~\cite{liang2020we}                                                       
&9.0 &10.8   &12.5  &15.1  &18.3  &13.1
&23.9 &25.2  &26.5  &28.4 &30.8   &27.0 
&33.7   &40.8 &46.9 &56.0   &66.1  &48.7
\\
TENT~\cite{wang2020tent}                                    &9.1 &10.7   &12.5  &15.2  &18.3  &13.2
&24.1 &25.4  &26.8  &28.7 &31.2   &27.2 
&33.2   &40.2 &46.0 &55.0   &65.3  &47.9
\\
SAR~\cite{niu2023towards}                                                       
&9.5 &11.6   &13.5  &16.6  &20.3  &14.3
&25.7 &27.0  &28.6  &30.8 &33.6   &29.1 
&33.3   &40.3 &46.1 &55.1   &65.2  &48.0\\ 
AdaContrast~\cite{chen2022contrastive}                                                       
&9.8 &11.9   &13.8  &17.0  &20.8  &14.7
&25.3 &26.8  &28.4  &30.9 &34.1   &29.1 
&35.2   &42.9 &49.4 &59.2   &69.2  &51.2
\\
CPL~\cite{conjugatepl}                                                     
&8.9 &10.5   &12.1  &14.6  &17.7  &12.8
&23.8 &25.0  &26.4  &28.3 &30.6   &26.8 
&32.7   &39.5 &45.6 &55.2   &65.9  &47.8
\\ 
Cotta~\cite{wang2022continual}                                                      
&9.3 &11.3   &13.0  &15.4  &18.4  &13.5
&26.3 &27.7  &29.4  &31.6 &34.5   &29.9 
&33.9   &41.6 &48.0 &57.7   &68.2  &49.9
\\
Rotta~\cite{yuan2023robust}                                                    
&9.7 &11.8   &13.8  &17.2  &21.7  &14.8
&30.1 &32.5  &35.0  &38.7 &42.8   &35.8 
&33.9   &42.0 &49.4 &59.5   &69.6  &50.9
\\ 
TSD~\cite{wang2023feature}                                                       
&9.5 &11.6   &13.5  &16.6  &20.3  &14.3
&26.6 &28.1  &29.9  &32.0 &35.5   &30.4 
&33.9   &41.5 &48.0 &57.8   &68.3  &49.9
\\
RMT~\cite{dobler2023robust}                                                     
&8.9 &10.3   &11.6  &\textbf{13.8}  &\textbf{16.5}  &\underline{12.2}
&24.8 &26.1  &27.6  &29.5 &31.8   &28.0 
&48.8   &53.0 &56.3 &61.8   &68.2  &57.6
\\ 
Ours                                                       
&\textbf{8.6} &\textbf{10.0}   &\textbf{11.5}  &\textbf{13.8}  &\underline{16.7}  &\textbf{12.1}
&\textbf{23.4} &\textbf{24.6}  &\underline{26.0}  &\textbf{27.8} &\textbf{30.0}   & \textbf{26.4}
&33.2   &40.1 &45.8 &54.8   &64.7  &47.7
\\
\midrule

SHOTIM~\cite{liang2020we}
&9.0 &10.7   &12.4  &15.1  &18.2  &13.1
&23.9 &25.2  &26.5  &28.4 &30.8   &27.0 
&33.7   &40.9 &47.0 &56.1   &66.2  &48.8      
\\
+ Ours
&\textbf{8.6} &\textbf{10.0}   &\underline{11.6}  &\underline{13.9}  &17.3  &12.3
&24.0   &\textbf{24.6} &\textbf{25.9} &28.2   &\underline{30.1}  &\underline{26.6}   
&33.2  &38.9 &43.3 &50.5   &59.3  &45.0
         \\ 
         \midrule
EATA~\cite{niu2022efficient}
&9.0 &10.7  &12.3  &15.0 &18.1   &13.0
&24.2 &25.6  &27.1  &29.2 &31.6   &27.5 
&32.1   &38.1 &43.1 &50.9   &60.5  &44.9      
\\
+ Ours
&\textbf{8.6} &\underline{10.1}   &\underline{11.6}  &14.0  &16.9  &\underline{12.2}
&23.8 &25.0  &26.4  &28.1 &30.6   &26.8 
&\textbf{31.9}   &\underline{37.3} &41.8 &49.1   &58.0  &43.6  
\\ 
\midrule
ROID~\cite{marsden2023universal}
&8.9 &10.5   &12.0  &14.6  &17.8  &12.8
&23.6 &24.8  &26.1  &28.1 &30.4   &\underline{26.6} 
&31.8   &\textbf{37.0} &\underline{41.3}  &\underline{48.5}   &\underline{57.3}  &\underline{43.2}      
\\
+ Ours
&\underline{8.7} &10.3   &11.8  &14.3  &16.9  &12.4
&\underline{23.5} &\underline{24.7} &\underline{26.0}  &\underline{27.9}   &\underline{30.1}   &\textbf{26.4} 
&\underline{32.0}   &\textbf{37.0} &\textbf{41.1} &\textbf{48.0}   &\textbf{56.2}  &\textbf{42.9}   
\\

\bottomrule
\end{tabular}
\label{all5corruption}
        
\end{table*}

\textbf{Domain generalization.}
We compare our approach with state-of-the-art methods on the four DG benchmarks and report the results in Table~\ref{table:pacs}. ``Source'' represents the source model's performance without any adaptations.

Table~\ref{table:pacs} shows that most of the TTA methods improve the source model's performance on
the PACS and VLCS datasets. However, they degrade model performance on more challenging datasets like OfficeHome and TerraIncognita. This is because the source model's performance on these databases is quite poor. Furthermore, it is more likely to generate incorrect pseudo-labels, impairing the existing TTA methods' robustness.

In contrast, as demonstrated in Table~\ref{table:pacs}, our approach significantly outperforms existing TTA methods across all four datasets. Specifically, our approach outperforms the source models on PACS, VLCS, OfficeHome, and TerraIncognita by 7.5\%, 5.8\%, 1.6\%, and 4.6\%, respectively. Also, it achieves the highest average accuracy across the four DG benchmarks, reaching 73.3\%. Moreover, compared with the second-best method~\cite{choi2022improving}, our approach is easier to use, as it is free from auxiliary image synthesis tasks.

\textbf{Single domain generalization.} 
We evaluate our method's performance on a significantly more challenging task (\textit{i.e.}, SDG). Compared to the traditional domain generalization task, SDG utilizes single-domain data for source model training, which means the training data lacks style variations.
This poses a significant challenge for self-training-based methods, as the source model that overfits a single domain may yield more unreliable pseudo-labels.

As shown in Table~\ref{domainnet-126}, the source models' average classification error rate is 45.3\%. 
Most self-training-based methods~\cite{wang2020tent,lee2013pseudo,conjugatepl} increase model performance. Specifically, our approach reduces the source model's average classification error rate from 45.3\% to 38.4\%.

Morover, Cotta~\cite{wang2022continual} and Adacontrast~\cite{chen2022contrastive} employ strong data augmentation techniques during TTA, and their performance is similar to DPL. 
Then, we combined DPL with Adacontrast; the results show that combinning them further improves TTA performance, achieving an 8.0\% reduction in the average classification error rate compared to the source model.

\newcommand{\tabincell}[2]{\begin{tabular}{@{}#1@{}}#2\end{tabular}}
\begin{table*}[t]
\caption{Ablation study on our methods' key components.}
\vspace{-0.25cm}
\centering
\begin{center}
\begin{tabular}
{p{2.5cm}<{\centering}| p{0.6cm}<{\centering} p{0.6cm}<{\centering} p{0.3cm}<{\centering} p{0.80cm}<{\centering}|c|c|c|c|c}
\hline
   & \multicolumn{4}{c|}{Components} &\multicolumn{5}{c}{Datasets}  \\
   
  \cline{2-10}
    Methods  &DPL$^{o}$ &DPL$^{*}$ & DPL &DPL${^\dag}$  &PACS  &VLCS   &OfficeHome  &TerraIncognita  &AVG\\
  \hline
  \hline
   \multirow{2}*{\tabincell{c}{Source\\Baseline}}
 &-  &- &-                    &-          &82.4 &74.4 &67.7  &47.5   &68.4 \\
      &-  &- &-   &-          
  &82.1 &74.4   &68.3  &49.6   &68.6\\
  \hline
  \multirow{3}*{\tabincell{c}{Single-Loss}}
&\checkmark  &- &-                   &-          &89.1 &78.8 &68.3 &49.9 &71.6\\
  &-  &\checkmark &-                   &-         &88.7&79.4&68.8&50.0&71.7\\
  ~   &-  &- &\checkmark   &-             &89.4&81.1&69.5&50.9&72.7\\
  \hline
  \multirow{3}*{\tabincell{c}{Combinations}}
  &\checkmark  &- &-    &\checkmark          &89.3 &79.3  &68.5  &52.0 &72.3 
  \\
  
  &-  &\checkmark &-    &\checkmark     
  &89.2	&81.4	&69.2	&51.7	&72.9\\
  &-  &- &\checkmark    &\checkmark          &89.9	&81.8	&69.3	&52.1	&73.3
  \\
  \hline
\end{tabular}
\end{center}
\label{tab:component}
\end{table*}

\textbf{Image corruption.} 
We provide the classification error rate for three widely-used image corruption benchmarks: CIFAR-10-C, CIFAR-100-C, and ImageNet-C. 
For ImageNet-C, we present detailed results for 15 corruption types at the most challenging severity level 5 in Table~\ref{tta.reset.imagenetc}. To further demonstrate our approach's effectiveness and flexibility, we evaluate its performance on all five corruption levels and summarize the experimental results in Table~\ref{all5corruption}.

In these tables, ``SOURCE'' represents the source model's performance without any adaptations. The poor results suggest a high vulnerability to corrupted images.
For the highest corruption level,
we observe that our approach outperforms the self-training-based methods that adopts a single entropy minimization or CE loss functions, such as PL~\cite{lee2013pseudo}, TENT~\cite{wang2020tent}, and CPL~\cite{conjugatepl}, by a large margins across the three datasets. Specifically, our approach achieves an average error rate of 16.7\%, 30.0\%, and 64.7\% on CIFAR-10-C, CIFAR-100-C, and ImageNet-C, respectively. Moreover, we discover that our approach can be combined with many existing loss functions. When equipped with our approach, SHOTIM~\cite{liang2020we}, EATA~\cite{niu2022efficient}, and ROID's~\cite{marsden2023universal} average error rates decrease by 0.9\%, 1.2\%, and 0.9\% on CIFAR-10-C, 0.7\%, 1.0\%, and 0.3\% on CIFAR-100-C, and 6.9\%, 2.5\%, and 1.1\% on ImageNet-C.

Furthermore, Table~\ref{all5corruption} demonstrates that our approach significantly reduces the source model's average error rate for CIFAR-10-C, CIFAR-100-C, and ImageNet-C across five corruption levels by 14.5\%, 8.5\%, and 13.1\%, respectively. Moreover, when combined with SHOTIM~\cite{liang2020we}, EATA~\cite{niu2022efficient}, and ROID~\cite{marsden2023universal}, our approach effectively reduces their average error rate by 0.8\%, 0.8\%, and 0.4\% on CIFAR-10-C, 0.4\%, 0.7\%, and 0.2\% on CIFAR-100-C, and 3.8\%, 1.3\%, and 0.3\% on ImageNet-C.

\subsection{The Ablation Study}

We conduct ablation studies on the four DG benchmarks. We use PL~\cite{lee2013pseudo} as the baseline, which updates the full-model parameters with the CE loss according to the samples with confident pseudo-labels in the target data. We run each experiment with three seeds and report the average performance following~\cite{wang2023feature}.

\textbf{Each key component's effectiveness.}
Table~\ref{tab:component} displays the performances of the proposed methods: DPL$^{o}$ (Eq.~\ref{eq.3}), DPL$^{*}$ (Eq.~\ref{eq.4}), and DPL (Eq.~\ref{eq.7}). Each method can perform TTA independently. We also present the DPL's performance when it utilizes samples with unconfident pseudo-labels (\textit{i.e.}, DPL$^{\dag}$).

We observe that the DPL variants significantly outperform the baseline model on the four benchmarks, achieving a 3\%, 3.1\%, and 4.1\% improved performance on average for DPL$^{o}$, DPL$^{*}$, and DPL, respectively. 
These results validate our proposed prototype-centric computation strategy's effectiveness. Besides, DPL outperforms DPL$^{o}$ and DPL$^{*}$ by a 1\% margin on average, demonstrating our consistency regularization method's effectiveness in Eq.~\ref{eq.6}.
Furthermore, utilizing the samples with unconfident pseudo-labels further improves the average performance of DPL$^{o}$, DPL$^{*}$, and DPL by 0.7\%, 1.2\%, and 0.6\%, respectively. These results demonstrate DPL's ability to leverage samples with unconfident pseudo-labels.

\begin{figure*}
    \centering
    \includegraphics[width=0.9\textwidth]{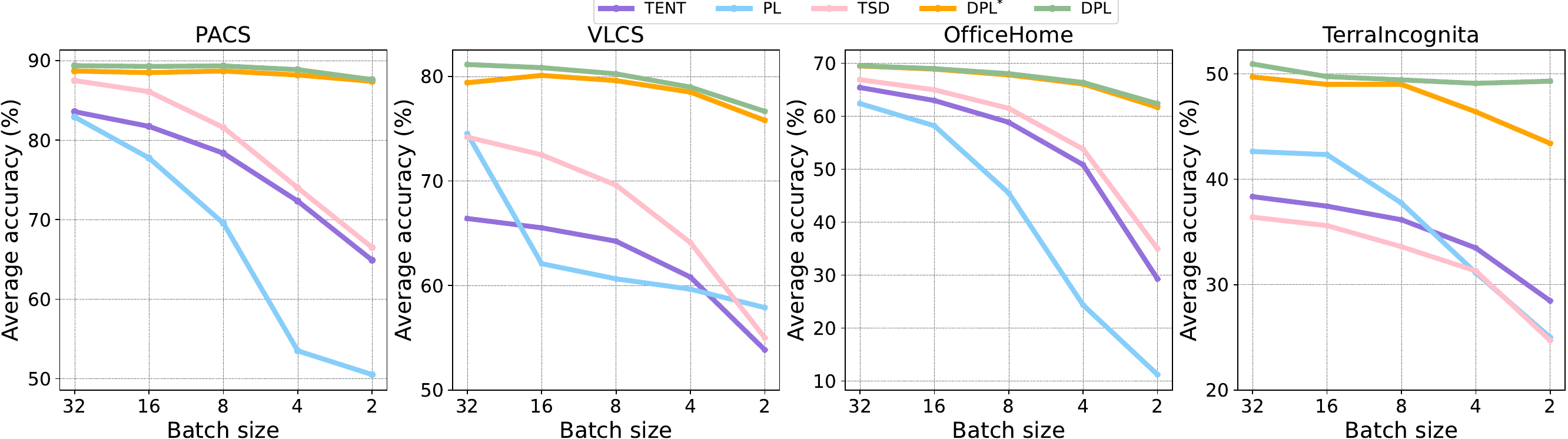}
    \caption{Performance comparison between TENT~\cite{wang2020tent}, PL~\cite{lee2013pseudo},  TSD~\cite{wang2023feature}, DPL$^{*}$, and DPL 
    on the four DG benchmarks with different batch sizes.}
    \label{ablation-bs}
\end{figure*}

\textbf{The method's robustness for small batch sizes.} 
In real-world applications, the amount of target domain data in each batch is usually unpredictable. Therefore, it is necessary to test the TTA methods' performance with various batch sizes. Fig.~\ref{ablation-bs} illustrates the experimental results and those of the three representative methods: PL~\cite{lee2013pseudo}, TENT~\cite{wang2020tent}, and TSD~\cite{wang2023feature}. They adopt the popular entropy minimization or cross-entropy loss for model optimization.

As shown in Fig.~\ref{ablation-bs},  PL~\cite{lee2013pseudo}, TENT~\cite{wang2020tent}, and TSD~\cite{wang2023feature} performance levels drop significantly as the batch size decreases. This may be because a small batch size leads to a more significant label noise impact on model optimization.
In contrast, the performance of DPL$^{*}$ and DPL is considerably more robust for small batch sizes. This is because each prototype is optimized in a disentangled manner, reducing the impact of label noise on class prototype optimization.

Moreover, the regularization term (Eq.~\ref{eq.6}) in DPL further enhances the robustness of DPL$^{*}$, especially for the TerraIncognita database~\cite{beery2018recognition} where the source model's performance is quite low. The above results justify our methods' robustness for small batch sizes.

\textbf{Sensitivity study to hyper-parameters.}
We evaluate DPL's sensitivity to three hyper-parameters, including the temperature $\tau$, the momentum coefficient $\eta$, and the trade-off parameter $\beta$. 
We also assess DPL and PL's performances when updating different model parameters (\textit{i.e.}, BN, the classifier, the feature extractor, and the full model, respectively). 

\begin{figure*}
    \centering
    \includegraphics[width=0.9\textwidth]{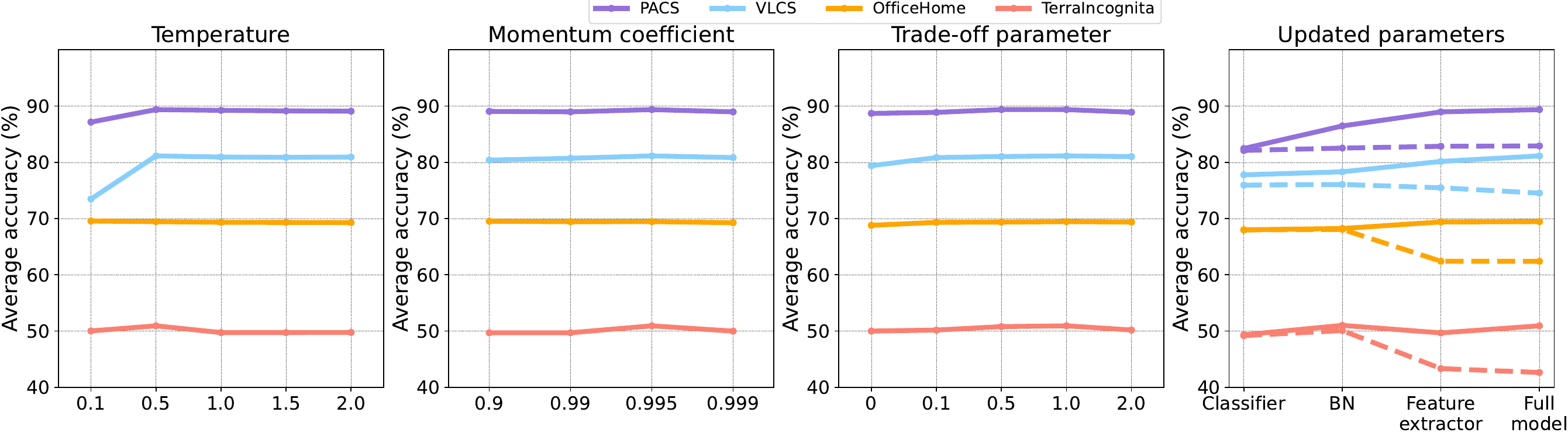}
    \caption{Performance of DPL with different hyper-parameter values, including (1) the temperature $\tau$, (2) the momentum coefficient $\eta$, (3) the trade-off parameter $\beta$, and (4) the range of updated parameters. The solid and dashed lines in (d) stand for the performance of DPL and PL, respectively.}
    \label{ablation-sen}
\end{figure*}

The experimental results are summarized in Fig.~\ref{ablation-sen}, which shows that the performance of DPL is robust for all three hyper-parameters. 
Moreover, we observe that updating more model parameters results in a severe decline in PL's performance level due to the erroneous gradients affecting more model parameters. In contrast, DPL benefits from the prototype-centric computation mode, producing more robust gradients. Hence, it achieves the best performance when all the model parameters are updated.

\begin{table}[htp]
\setlength\tabcolsep{8pt}
\renewcommand{\arraystretch}{0.8}
\begin{center}
		\caption{Performance comparisons of DPL with different backbones on the PACS database~\cite{li2017deeper}. $\Delta$(\textcolor{red}{$\uparrow$}) denotes the performance improvement level. The \textbf{best} and the \underline{second-best} methods are highlighted.}
		\vspace{-0.1cm}
		\resizebox{\linewidth}{!}{
			\begin{tabular}{l|cccc|c|c}
			\toprule
 Backbones & A & C & P & S & Average &$\Delta$(\textcolor{red}{$\uparrow$}) \\ \midrule \midrule
           
  {\text{ResNet-18~\cite{he2016deep}}} &78.7	&76.2	&94.9	&70.7	&80.1  & - \\ 
  \text{+PL~\cite{lee2013pseudo}} &79.7	&76.1	&95.0	&71.5	&80.6 & \textcolor{black}{0.5}
  \\
\text{+T3A~\cite{iwasawa2021test}} &80.2	&79.4	&\underline{96.2}	&72.2	&82.0 & \textcolor{black}{1.9}
\\
\text{+TSD~\cite{wang2023feature}} &\underline{85.3}	&\textbf{84.3}	&96.0	&\underline{74.0}	&\underline{84.9} &\textcolor{black}{\underline{4.8}}
  \\
 {\text{+DPL}} &\textbf{85.4}	&\underline{82.6}	&\textbf{96.9}	&\textbf{80.8}	&\textbf{86.4} &\textcolor{black}{\textbf{6.3}}  \\
\midrule
{\text{ViT-B/16}~\cite{dosovitskiy2020image}} &88.2	&85.0	&97.7	&78.0	&87.2  &-\\ 
 \text{+PL~\cite{lee2013pseudo}} &88.5	&\underline{88.3}	&97.8	&78.1	&88.2 &\textcolor{black}{1.0}
  \\
\text{+T3A~\cite{iwasawa2021test}} &88.6	&87.1	&\underline{98.6}	&79.1	&88.4 &\textcolor{black}{1.2}
\\
\text{+TSD~\cite{wang2023feature}} &\underline{90.9}	&86.9	&98.1	&\underline{79.2}	&\underline{88.8} &\textcolor{black}{\underline{1.6}}
  \\
{\text{+DPL}} &\textbf{92.5}	&\textbf{90.9}	&\textbf{98.7}	&\textbf{81.4}	& \textbf{90.9} &\textcolor{black}{\textbf{3.7}}  \\
            \midrule
\text{Mixer-B/16~\cite{tolstikhin2021mlp}} &81.4	&71.3	&95.9	&60.9	&77.4 &-\\
 \text{+PL~\cite{lee2013pseudo}} &80.3	&73.3	&95.7	&\underline{63.2}	&78.1 &\textcolor{black}{0.7}
  \\
\text{+T3A~\cite{iwasawa2021test}} &\underline{81.9}	&\underline{74.2}	&\underline{96.1}	&63.0	&\underline{78.8} &\textcolor{black}{\underline{1.4}}
\\
\text{+TSD~\cite{wang2023feature}} &81.3	&71.9	&\textbf{96.3}	&62.8	&78.1 &\textcolor{black}{0.7}
  \\
{\text{+DPL}} &\textbf{84.2}	&\textbf{82.5}	&96.0	&\textbf{73.3}	&\textbf{84.0} &\textcolor{black}{\textbf{6.6}}  \\
            \bottomrule 
            \end{tabular}\label{ablation-scale}
            }
	\end{center}
\end{table}


\textbf{Application to various backbones.} 
In this experiment, we validate the effectiveness of DPL$^{\dag}$ with various backbones, including ResNet-18~\cite{he2016deep}, ViT-B/16~\cite{dosovitskiy2020image} and Mixer-B/16~\cite{tolstikhin2021mlp}.
As the latter two backbones do not adopt BN layers, we exclude the BN calibration-based methods from the comparison.
The performance of the pre-trained source model, existing representative TTA methods~\cite{lee2013pseudo,iwasawa2021test,wang2023feature}, and DPL are summarized in Table~\ref{ablation-scale}.

The table shows that the compared methods'~\cite{lee2013pseudo, iwasawa2021test,wang2023feature} performance levels slightly increase for the three backbones. 
In contrast, DPL's performance level is consistent and significantly increases by 6.3\%, 3.7\%, and 6.6\% for ResNet-18, ViT-B/16 and Mixer-B/16, respectively. Particularly, DPL's performance substantially increases by 10.1\%, 3.4\% and 12.4\% for ResNet-18, ViT-B/16 and Mixer-B/16, respectively, on the most challenging ``sketch'' domain.
The above experimental results validate the effectiveness and flexibility of DPL for the TTA task.

\begin{table}[htp]
\setlength\tabcolsep{8pt}
\renewcommand{\arraystretch}{0.8}
\begin{center}
		\caption{Performance comparisons with two variants in sampling $\textbf{x}^b$.}
		\resizebox{\linewidth}{!}{
			\begin{tabular}{l|cccc|c}
			\toprule
  & PACS & VLCS & OfficeHome & TI & Average \\ \midrule \midrule
           
  {\text{Confident ones}} &88.7  &80.9  &69.4  &51.2  &72.6 \\ 
\midrule
{\text{All samples}} &88.8  &81.3  &69.5  &51.6  &72.8  \\ 
            \midrule
\text{DPL$^{\dag}$} &\textbf{89.9}	&\textbf{81.8}	&\textbf{69.3}	&\textbf{52.1}	&\textbf{73.3}  \\
            \bottomrule 
            \end{tabular} \label{ablation-cr} 
            }
	\end{center}
\end{table}

\textbf{Comparisons in variants in sampling $\textbf{x}^b$.} Two natural variants include randomly selecting $\textbf{x}^b$ from all the samples in a batch or only those with confident pseudo-labels. We compare their performance with our described strategy in Section~\ref{sec:methods}. 

\begin{table}[htp]
\setlength\tabcolsep{8pt}
\renewcommand{\arraystretch}{0.8}
\begin{center}
		\vskip 0.10in
		\caption{Performance comparisons with representative prototype-based loss functions.}
		\resizebox{\linewidth}{!}{
			\begin{tabular}{l|cccc|c}
			\toprule
             Method  & PACS & VLCS & OfficeHome & TI & Average \\ \midrule \midrule
            \text{PL(CE)} \multirow{8}{*} &82.1 &74.4  &68.3  &49.6  &68.6  \\
            \midrule
            \text{Arcface~\cite{arcface}} \multirow{8}{*} &84.0 &72.4 &68.5  &49.2  &68.5
            \\  
            \text{Proxy-NCA~\cite{movshovitz2017no}} &83.4  &75.4  &68.4  &48.5  &68.9  \\
            \midrule
            {DPL} &\textbf{89.4}&\textbf{81.1}&\textbf{69.5}&\textbf{50.9}&\textbf{72.7} \\ 
            \bottomrule 
            \end{tabular} \label{table:prototype} 
            }
	\end{center} 
\end{table}

As shown in Table~\ref{ablation-cr}, both variants' performances are poorer than our methods. This indicates that the samples with unconfident pseudo-labels are more beneficial, as they usually provide more target domain information that benefits the TTA task.

\textbf{Comparisons with prototype-based loss functions.} We compare DPL's performance with two representative prototype-based loss functions (~\textit{i.e.}, Arcface\cite{arcface} and Proxy-NCA \cite{movshovitz2017no}) in Table~\ref{table:prototype}. The angular margin in Arcface is set to zero.

As shown in Table~\ref{table:prototype}, the 
Arcface and Proxy-NCA's performances are similar to that of PL (\textit{i.e.}, the original CE loss). This is because they are all designed using sample-centric computation, which is vulnerable to label noise. In comparison, DPL performs better than Arcface and Proxy-NCA because it decouples prototype learning. This further demonstrates that the decoupled optimization of class prototypes facilitates reliable TTA.

\textbf{Comparisons in a supervised setting.} We compare the DPL's performance with the CE loss in a supervised setting. We conduct experiments with the standard CIFAR-10 database~\cite{krizhevsky2009learning} using ResNet-50~\cite{he2016deep} as the backbone and report the classification accuracy according to the official evaluation protocol~\cite{krizhevsky2009learning}.

The classification accuracies are 95.3$\%$ and 78.7$\%$ for CE and DPL, respectively. 
This means that fitting each sample's label accurately (\textit{i.e.}, CE loss) is beneficial in a supervised setting. DPL's performance is worse than the CE loss in the supervised setting, likely due to its decoupling of class prototype optimization, which weakens the model's discriminative power when accurate labels are available. However, as shown in Table~\ref{table:prototype}, DPL outperforms CE by an average of 4.1\% in the TTA setting. This indicates that DPL is dedicated to the TTA task.


\section{Conclusion}
\label{sec:conclusion}
\vspace{-0.15cm}
In this paper, we propose a novel DPL approach for reliable TTA. 
We demonstrate that the prototype-centric loss computation mode is more suitable than the sample-centric mode during TTA. When dealing with the noisy pseudo-labels, our DPL method's robust adaptation of model parameters under different domain shifts displays improved performance. 
Additionally, we incorporate a memory-based strategy that enhances DPL method's robustness, ensuring its effectiveness even with limited batch sizes.
Moreover, we introduce a consistency regularization-based method to leverage samples with unconfident pseudo-labels more effectively.
Finally, our method significantly outperforms state-of-the-art TTA methods on DG benchmarks and reliably enhances the performance of self-training-based methods on image corruption benchmarks.

\bibliographystyle{IEEEtran}
\bibliography{egbib}

\end{document}